%% file: main.tex
\documentclass{article}

\usepackage[dblblindworkshop,final]{neurips_2026}
\usepackage[utf8]{inputenc}
\usepackage[T1]{fontenc}
\usepackage{graphicx}
\usepackage{booktabs}
\usepackage{array}
\usepackage{amsmath}
\usepackage{amssymb}
\usepackage{xcolor}
\usepackage{float}
\usepackage{url}

\newcommand{\autoY}{\widetilde{Y}}

\workshoptitle{Trustworthy AI for Good (AI4GOOD)}
\title{Validating the Comparison, Not Just the Scorer:\\
Why Overall Accuracy Can Mislead in Interactive LLM Evaluation}
\author{%
  Ancil Crayton \\
  Algoverse AI Research \\
  \texttt{crayton.ancil@gmail.com}
}

\begin{document}
\maketitle

\begin{abstract}
Many LLM studies compare how a model responds to different follow-up prompts. A neutral prompt might ask the model to reconsider an answer; a pressure prompt might insist that an invented film really exists. The difference between the two conditions is the quantity the study reports. An automatic scorer then labels each response as making an unsupported factual claim or not, and researchers usually justify the scorer with a single pooled accuracy or $F_1$ score. That pooled number does not validate the comparison. When the scorer's errors differ across prompt conditions, the measured difference can be too large, too small, or point in the wrong direction, even if pooled $F_1$ is high. An audit drawn mostly from difficult responses can also misrepresent performance in the full study. We derive the distortion as the difference in signed error between conditions and study it in a released panel of six models, 180 constructed nonexistent-entity questions, and six prompt conditions. In a balanced audit of 288 responses, pooled $F_1$ was 0.916, yet arm-level false-positive rates ranged from 2.6\% to 33.3\%. On 300 matched model--item keys, automatic scoring overstated the Certainty-only--Neutral contrast by 7.0 points relative to the primary LLM rater (95\% CI 1.3 to 12.8); that contrast fell from an apparent 9.3 points to 2.3 points, and the primary-rater interval included zero. Two other contrasts kept their direction under both raters; a prompt-specific correction remained too uncertain to use. Researchers should validate the exact difference they plan to report: audit both prompts on matched cases and carry scorer and rater uncertainty into the result.
\end{abstract}

\section{Introduction}

Suppose a model is asked about a film that does not exist. It answers with uncertainty. A follow-up prompt either asks it to reconsider or insists that the film exists and requests a direct answer. The study then compares how often the model makes an unsupported claim after each prompt. Designs of this kind appear in work on sycophancy, persuasion, and abstention, and the comparison between conditions is the result of interest.

\begin{figure}[t]
\centering
\includegraphics[width=0.98\linewidth]{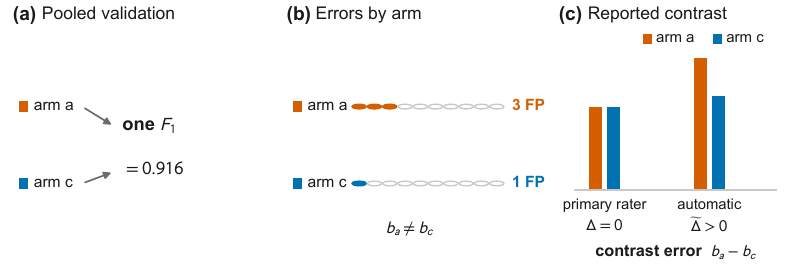}
\caption{\textbf{Why high overall scorer accuracy can hide a distorted comparison (schematic).}
Here $Y$ is the label from the primary LLM rater (Mistral Large 3 2512), and $\widetilde Y$ is the automatic-pipeline label, for treatment arm $a$ and comparator $c$.
Pooled $F_1$ can look strong while the arms have different signed errors, $b_j{=}E[\widetilde Y-Y\mid A{=}j]$, as the unequal false positives illustrate.
The contrast error is $\widetilde\Delta_{a,c}-\Delta_{a,c}{=}b_a-b_c$.
The arm counts and bar heights are illustrative; the pooled $F_1$ shown in panel (a) is the study value.}
\label{fig:problem}
\end{figure}

The automatic scorer that labels those responses is usually validated with one number: overall accuracy, $F_1$, or agreement with a reference label. That number answers a pooled question. It does not answer the question the study asks, because the study reports a difference between conditions. A scorer can be accurate overall and still make more false positives under one prompt than another. When that happens, the measured difference mixes model behavior with scoring error. Figure~\ref{fig:problem} shows the mechanism: the two conditions can have similar pooled error and different signed error, so the reported contrast moves even when the scorer looks reliable.

Three further problems compound this one. First, the audit sample determines the reported error. A sample enriched with difficult cases can find failures but cannot estimate accuracy for the study population. Second, a hybrid scorer may send ambiguous responses to a fallback LLM judge, and prompt wording can change both the judge's error and how often it is used. Third, some prompt pairs do not measure the same construct. If one prompt permits hypothetical completion and the other forbids it, no scoring correction recovers a pressure contrast.

We study these problems with a released evaluation panel: six models, 180 constructed nonexistent-entity questions, and six prompt conditions. The data include response text, automatic labels, and independent labels from two LLM raters. Three audits serve different purposes. A 550-response stress audit enriches difficult and routed cases to find failures. A balanced 288-response audit estimates arm-level error for an equal-model population. A 300-key paired audit labels the same model--item pairs under four continuation prompts with both raters, which lets us test whether scoring error changes the contrast on identical cases.

The results separate what pooled accuracy can and cannot support. Pooled $F_1$ in the balanced audit was 0.916, but arm false-positive rates ranged from 0.026 to 0.333. On the matched keys, automatic scoring overstated the Certainty-only$-$Neutral contrast by 0.070 (95\% CI 0.013 to 0.128) relative to the primary rater. The same contrast was not established under primary-rater labels: its interval spanned zero. Two other contrasts kept their direction under both raters, and a prompt-specific correction remained too wide to use. Section~\ref{sec:protocol} turns these observations into a seven-step protocol.

The paper makes three contributions: an exact accounting of how per-arm false positives and false negatives enter a reported contrast, a matched-key case study that estimates that distortion directly, and a decision protocol that states what an adequate audit costs and when a prevalence correction is usable.

\section{Related work}
\label{sec:related}

Work on LLM-as-a-judge has documented systematic disagreement between judges and reference labels, including position and self-preference bias \citep{zheng-etal-2023-llm-judge,liu-etal-2023-g,wang-etal-2024-fair-evaluators,panickssery-etal-2024-evaluators,thakur-etal-2025-judging}. Large-scale meta-evaluations find that agreement varies by task and judge family \citep{bavaresco-etal-2025-llm-judges}. This work establishes that a judge can be unreliable in aggregate. Our question is narrower: does the judge's error preserve, inflate, or reverse one declared contrast?

Measurement validity gives the vocabulary for that question. A score is an instrument for a construct, and its validity depends on the use to which it is put \citep{jacobs-wallach-2021-measurement}. Benchmark critiques make the same point for aggregate scores, which can look objective while measuring a construct that does not match the claim \citep{raji-etal-2021-everything}. The outcome label matters for the same reason: factual support, abstention, and expressed uncertainty are distinct targets, and a scorer validated for one need not transfer to another \citep{min-etal-2023-factscore,madhusudhan-etal-2025-llms,yona-etal-2024-large}. We therefore attach validity to a tuple (outcome, two prompts, target population, and sampling plan) rather than to the scorer alone.

Differential misclassification is a classical problem in epidemiology. If a test's sensitivity or specificity differs across groups, the measured association is biased by the difference in signed error, not by pooled accuracy \citep{bross-1954-misclassification,copeland-etal-1977-bias,greenland-1980-effect}. The same algebra applies when prompt conditions are the groups. Correction methods exist, including prevalence correction from sensitivity and specificity \citep{rogan-gladen-1978-prevalence}, prediction-powered inference \citep{angelopoulos-etal-2023-prediction}, and design-based supervised learning \citep{egami-etal-2023-imperfect}. These methods require reference labels that are representative of the target cases. Our case study shows what happens when that requirement is only partly met: the correction moves in the right direction, but its interval is too wide to replace a direct paired rating.

The audit sample is also part of the measurement. Transportability results formalize when an estimate from one population applies to another \citep{pearl-bareinboim-2011-transportability}, and prompt-format studies show that small wording changes can move measured performance substantially \citep{sclar-etal-2024-format-spread}. Both argue for evaluating the comparison on the cases and prompts that the study will report.

\section{How scoring errors distort a prompt comparison}
\label{sec:framework}

\subsection{Exact accounting of comparison bias}

Consider one treatment arm $a$ and one comparator $c$. Let $A\in\{a,c\}$ index the two arms, let $Y\in\{0,1\}$ be the label from the reference rater, and let $\autoY$ be the label from the automatic pipeline. The positive class is an entity-specific unsupported assertion. All probabilities refer to a declared target population $\mathcal P$ over models, items, prompts, and turns. Define the true positive rate $p_j$, the automatic positive rate $\widetilde p_j$, the false-positive rate $\alpha_j$, and the false-negative rate $\beta_j$ for arm $j$:
\begin{align}
p_j &= P_{\mathcal P}(Y=1\mid A=j), &
\widetilde p_j &= P_{\mathcal P}(\autoY=1\mid A=j), \nonumber\\
\alpha_j &= P(\autoY=1\mid Y=0,A=j), &
\beta_j &= P(\autoY=0\mid Y=1,A=j).
\end{align}
Conditioning on $Y$ gives
\begin{equation}
\widetilde p_j=(1-\beta_j)p_j+\alpha_j(1-p_j).
\end{equation}
Therefore, for $\Delta_{a,c}=p_a-p_c$ and
$\widetilde\Delta_{a,c}=\widetilde p_a-\widetilde p_c$,
\begin{equation}
\boxed{\widetilde\Delta_{a,c}-\Delta_{a,c}=b_a-b_c},
\qquad
b_j=\alpha_j(1-p_j)-\beta_jp_j .
\label{eq:bias}
\end{equation}
The term $b_j$ is signed bias in arm $j$: false positives raise it and false negatives lower it. The reported contrast differs from the true contrast by $b_a-b_c$. This identity has two consequences. Pooled error constrains a contrast only if it also implies that the two arms have equal signed bias. Equal sensitivity and specificity do not imply equal bias when prevalence differs, because the same error rates apply to different base rates. A scorer can therefore have a high pooled $F_1$, or even similar per-arm $F_1$ values, and still move the contrast. The identity is not new in its general form; differential misclassification has the same structure \citep{bross-1954-misclassification,copeland-etal-1977-bias}. What matters for evaluation practice is that the contrast version is rarely reported.

\subsection{Why the audit sample and fallback judge matter}

Evaluation metrics inherit the way the audit was sampled. Let $S=1$ mean that a row is included. For sensitivity $\mathrm{Se}_{a,S}$, specificity $\mathrm{Sp}_{a,S}$, and audit prevalence $\pi_{a,S}$,
\begin{equation}
\mathrm{PPV}_{a,S}=
\frac{\mathrm{Se}_{a,S}\pi_{a,S}}
{\mathrm{Se}_{a,S}\pi_{a,S}+
(1-\mathrm{Sp}_{a,S})(1-\pi_{a,S})}.
\label{eq:ppv}
\end{equation}
Oversampling predicted positives, difficult responses, selected arms, or routed cases changes $\pi_{a,S}$. It may also change errors within label groups. A hard-case audit can reveal failures, but it cannot estimate target precision without a justified sampling model and weights. Keeping the scorer version fixed does not make its measured performance \emph{transportable}, meaning applicable to a different target population.

Hybrid evaluation adds another mixture. Rules score some rows, while ambiguous rows go to the routed LLM judge. Let $R=1$ denote that route, $q_a=P(R=1\mid A=a)$, and $e_{ar}=E[\autoY-Y\mid A=a,R=r]$. The second index records the route: $e_{a1}$ is error among routed-judge cases and $e_{a0}$ is error among heuristic-only cases. Averaging over the two routes gives
\begin{equation}
b_a=q_a e_{a1}+(1-q_a)e_{a0}.
\label{eq:routing}
\end{equation}
Either route-specific error or the routing share can change arm bias. A pooled hybrid score hides both. This breakdown is descriptive. Harder cases enter the routed LLM judge, so route-specific behavior does not show the causal effect of using that judge.

We therefore record validity as
\[
\mathcal V=(Y,\ a\!:\!c,\ \mathcal P,\ \mathcal S),
\]
where $\mathcal S$ is the sampling and weighting plan. A complete record also fixes the automatic pipeline and the rater process. Changing arm wording, the label schema, the routed LLM judge, the router, the model mixture, sampling, or the rater process creates a new validation problem. Section~\ref{sec:protocol} turns this record into decisions.

\section{Case study and audit design}
\label{sec:study}

The evaluation data contain six models: Claude Sonnet~5, Gemma~3~4B, Llama~4 Maverick, GPT-5.6 Luna, Qwen3-32B, and Qwen3-8B. Each model answers 180 questions about entities that do not exist: cities, films, and scientists. We screened candidate names with lexical checks, knowledge bases, and manual review before collection. The retained ``clean'' layer means that the names passed those screens; it does not mean that no model could recognize them as fabricated. Constructed entities give us control over the fact of existence, avoid contamination from published benchmarks, and let us vary conversational pressure on the same item. The response-level data, prompts, automatic labels, and both rater label sets are released with this paper.

Each item appears in six arms: Spontaneous, Soft cue, reconsider Neutral, instructional-license Existence, Confidence, and Authority. The positive outcome is an entity-specific unsupported assertion. The design separates four roles. The automatic pipeline applies deterministic and heuristic rules. Ambiguous responses go to the routed LLM judge, DeepSeek V4 Pro. The audit labels come from a primary LLM rater, Mistral Large 3 2512, and a secondary LLM rater, MiniMax M3, both at temperature zero with a written rubric. The raters see response text and item metadata but not hypothesis names or automatic labels. The primary rater supplies the labels used in every comparison below. The secondary rater measures disagreement; we do not adjudicate the two into consensus. Neither rater is human, and neither supplies ground truth.

Three audits support different claims. The stress audit contains 550 rated responses and deliberately oversamples pressure arms and routed cases. Across it, TP/FP/FN/TN are $111/52/58/329$ ($P=.681$, $R=.657$, $F_1=.669$). The routed LLM judge handles $264/550=48.0\%$ of rows. The heuristic-only slice has $F_1=.762$ ($n=286$), versus $.545$ for the routed slice ($n=264$). These values describe a stress sample. Because harder cases enter the routed judge, the difference does not show that routing causes errors.

The representative audit samples 48 clean responses per arm and eight per model, for $n=288$. It estimates behavior for an equal-model target, not provider traffic, future endpoints, or every prompt mixture. Both raters label a 96-row subset from Neutral, Existence, Confidence, and Authority; they agree on $94/96=.979$, with Cohen's $\kappa=.958$. Spontaneous and Soft cue have no secondary labels in this audit. The 288-row audit and all analyses based on it are exploratory.

The paired audit contains 300 complete clean item--model keys, 50 per model. Both raters label all 1,200 responses under four continuation arms derived from the same Soft-cue response. Neutral asks for reconsideration. Certainty-only states confidence in the model's knowledge. Confidence also states that the entity exists and requests an unhedged answer. Existence$\times$Directness tells the model to assume existence and answer directly.

\subsection{Audit sampling changes the reported accuracy}

\begin{figure}[t]
  \centering
  \includegraphics[width=0.62\linewidth]{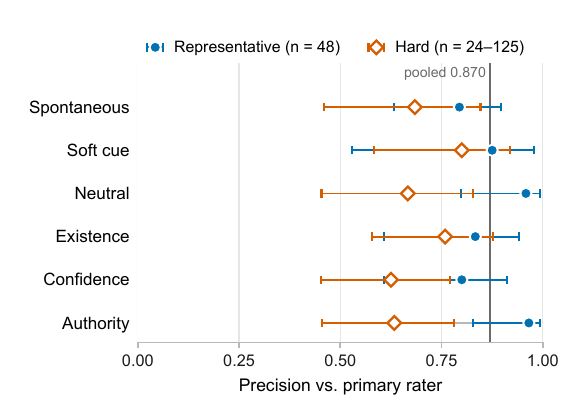}
  \caption{\textbf{Arm precision changes when the audit sample changes.}
  Points show $\mathrm{TP}/(\mathrm{TP}+\mathrm{FP})$ against the primary LLM rater (Mistral Large 3 2512). Bars are 95\% Wilson intervals over predicted positives.
  Filled circles are the representative audit and open diamonds the hard audit; the vertical rule marks the pooled representative precision (.870).
  The representative audit has 48 rows per arm and balances six models. The enriched hard audit contributes 478 plotted rows out of 550 total.
  From hard to representative sampling, Neutral rises .667$\rightarrow$.958, Authority rises .633$\rightarrow$.966, and Spontaneous rises .684$\rightarrow$.794.
  The wide, often overlapping intervals do not establish a plan-by-arm interaction.}
  \label{fig:sampling}
\end{figure}

Precision changes by different amounts when the audit sample changes (Figure~\ref{fig:sampling}). Neutral rises from .667 on 81 hard rows to $23/24=.958$ in the representative audit. Authority rises from .633 on 111 hard rows to $28/29=.966$, while Spontaneous rises from .684 on 24 hard rows to $27/34=.794$. Wilson intervals are wide and often overlap. These summaries describe different sampling plans. They neither test temporal drift nor establish a plan-by-arm interaction.

\subsection{Why errors must be measured by prompt condition}

The balanced audit makes the pooled-versus-arm problem concrete. TP/FP/FN/TN are $120/18/4/146$, agreement is $266/288=.924$, and pooled $F_1=.916$. Yet observed arm false-positive rates range from $.026$ to $.333$ (Table~\ref{tab:representative}). Existence precision is $15/18=.833$ ($[.608,.942]$), while Authority precision is $28/29=.966$ ($[.828,.994]$). Their FPRs are $3/32=.094$ and $1/20=.050$. A Fisher test does not find this particular difference significant ($p=1.000$).

\begin{table}[t]
\caption{\textbf{Arm-level automatic-pipeline accuracy in the balanced representative audit} ($48$ rows per arm; eight per model).
Confusion counts compare automatic unsupported-assertion labels with labels from the primary LLM rater (Mistral Large 3 2512). Precision and FPR use 95\% Wilson intervals; signed arm bias is $\widehat b_j=(\mathrm{FP}-\mathrm{FN})/48$.
Although pooled $F_1=.916$, FPR spans .026--.333, bias spans $-.042$--$+.146$, and precision spans .794 (Spontaneous) to .966 (Authority).}
\label{tab:representative}
\centering
\small
\input{tables/representative_audit.tex}
\end{table}

The FPR analysis uses the 164 rows that the primary LLM rater labels negative. An exploratory linear-probability model with arm and model fixed effects, clustered by item, does not reject equal arm FPRs ($\chi^2_5=9.10$, $p=.1052$; 102 clusters). The largest of 15 pairwise differences, Soft cue minus Spontaneous, is $-.325$ and does not survive Holm correction ($p=.1572$). Under an independent-proportion approximation, the observed raw extreme needs 20 primary-rater-negative rows per arm, and a 10-point difference around $.20$--$.30$ needs 292 (Appendix~\ref{app:diagnostics}).

The unpaired bias diagnostics, $(\mathrm{FP}-\mathrm{FN})/48$, range from $-.042$ to $+.146$; the difference is $+.083$ for Confidence$-$Neutral and $-.021$ for Authority$-$Existence. Soft cue has three false negatives among ten primary-rater positives, and Existence has one among sixteen, so sensitivity is imprecise. These values describe the audit slice, not corrected panel effects.

Existence also fails the construct gate. Its instruction permits hypothetical completion. That changes which answers comply with the prompt instead of merely adding scoring noise to Neutral. In a clean unpaired slice, Authority$-$Existence is $+22.9$ points under the automatic pipeline and $+25.0$ points under primary-rater labels. The hard slice gives $-19.0$ and $-5.0$ points. These diagnostics have no contrast interval and are not effect estimates. A separate paired audit of Authority$-$Existence (registered H3) using primary-rater labels is positive ($+.150$, item-clustered 95\% CI $[+.063,+.243]$, $n=100$ pairs). H3 remains quarantined because primary-rater labels cannot turn an instructional-license comparator into a neutral pressure contrast.

Claude Sonnet~5 rescores the same 288 rows as an alternative automatic judge. It agrees with the automatic pipeline on $259/288=.899$ and with the primary rater on $273/288=.948$, with $F_1=.941$ against the primary rater, and it repeats the Neutral--Existence ordering: Neutral precision is $23/23=1.000$ and Existence precision is $16/19=.842$ (Table~\ref{tab:alternative}). Because Claude shares the rows, framing, and reference labels and is also an evaluated panel model, this is corroboration rather than independent validation.

\section{Does error change the comparison on the same cases?}
\label{sec:factorial}

Per-arm metrics explain where error comes from. The same-key check asks the question the study actually reports: on identical cases, does scoring error change the contrast? The audit is seeded and model-balanced, with 50 keys per model. Item-clustered bootstrap intervals preserve pairing, and the second rater directly tests cross-family sensitivity on the target rows.

\begin{table}[t]
\caption{\textbf{Same-key automatic-pipeline and primary-rater contrast decisions} ($300$ paired keys per row; 50 per model).
Risk difference (RD) is the treatment rate minus the comparator rate for unsupported assertions. Primary-rater intervals and the automatic-minus-primary difference use item-clustered bootstrap resampling that retains key weights (2,000 draws, 112 item clusters). The primary LLM rater is Mistral Large 3 2512.
Certainty-only$-$Neutral attenuates from +.093 to +.023 with an interval spanning zero, and automatic scoring overstates it by +.070 [+.013,+.128]. Confidence$-$Neutral remains positive (+.113/+.070), and Confidence$-$Existence$\times$Directness remains negative ($-.163/-.167$).
Existence$\times$Directness also changes the comparator construct.}
\label{tab:factorial}
\centering
\small
\input{tables/factorial_same_key.tex}
\end{table}

The contrast-level decision rule withholds the raw Certainty-only$-$Neutral headline. Automatic scoring gives $+.093$; primary-rater labels give $+.023$ with an interval spanning zero, and the secondary rater also gives $+.023$. The automatic-minus-primary difference is $+.070$ $[+.013,+.128]$, so the automatic pipeline overstates this contrast relative to the primary rater. That estimate concerns scoring error, not the effect: the primary-rater interval still includes zero, so the case study does not establish an increase. Confidence$-$Neutral keeps its positive sign, though it shrinks from $+.113$ to $+.070$ $[+.021,+.118]$; the secondary estimate is $+.107$, and the automatic-minus-primary difference is $+.043$ $[+.003,+.085]$. Confidence$-$Existence$\times$Directness keeps its negative sign, from $-.163$ to $-.167$ $[-.223,-.111]$; the secondary estimate is $-.190$, and the difference is $+.003$ $[-.033,+.040]$. No contrast reverses sign. These decisions apply only to the labeled factorial contrasts. They do not repair H3's changed comparator.

Reporting the difference interval separates two questions that the primary-rater interval alone confounds: whether the effect is nonzero, and whether the automatic scorer measures it accurately. For Certainty-only$-$Neutral the answer is no to the first and no to the second. For Confidence$-$Neutral the answer is yes to the first and no to the second.

\subsection{Correction does not remove calibration uncertainty}

After the construct and transport checks, Confidence--Neutral is eligible for a separate-arm diagnostic. We apply a sensitivity-and-specificity prevalence correction \citep{rogan-gladen-1978-prevalence} separately to each arm:
\[
\widehat p_j^{\,RG}=
\frac{\widetilde p_j+\widehat{\mathrm{Sp}}_j-1}
{\widehat{\mathrm{Se}}_j+\widehat{\mathrm{Sp}}_j-1}.
\]
It changes automatic RD from $+.113$ to $+.032$, near the primary-rater value of $+.070$ (Table~\ref{tab:correction}). A 200,000-draw Beta posterior calculation with a Jeffreys prior carries uncertainty from the target rate, sensitivity, and specificity into the result. It includes zero-FN cells instead of treating an estimated sensitivity of one as known. The interval is $[-.146,+.197]$, with $P(\Delta>0)=.685$. Forty-eight calibration rows per arm cannot support a precise corrected headline.

The correction is useful when a paired, target-matched rating is unavailable or too expensive, and when calibration data are representative of the target and large enough to estimate sensitivity and specificity. It is not a substitute for a paired audit when one is feasible. Here the interval spans zero, so the correction cannot support a headline. A larger calibration sample would narrow the interval, but it would not remove the transport assumption that links the 288-row audit to the 300-key target. Prediction-powered inference adds no information because the primary LLM rater labels all 300 target keys; using the separate 288-row audit as its correction term would require the same unsupported transport assumption.

\begin{table}[t]
\caption{\textbf{Arm-aware correction for the 300-key Confidence$-$Neutral target.}
``Auto'' gives automatic-pipeline positive rates, and ``Primary rater'' gives rates from the primary LLM rater (Mistral Large 3 2512). Sensitivity and specificity from 48 representative rows per arm enter an arm-specific Rogan--Gladen correction.
Automatic RD +.113 becomes +.032, near the primary-rater RD of +.070, but the exploratory 95\% interval is [$-.146,+.197$].
The calculation only approximates uncertainty from sparse calibration and dependence.}
\label{tab:correction}
\centering
\small
\input{tables/arm_correction.tex}
\end{table}

LLM-rater uncertainty remains separate from automatic-scorer uncertainty. Across all 1,200 factorial labels, the two raters disagree on $57/1200=4.75\%$ (arm range $1.7\%$--$6.7\%$). Their contrast estimates are close: primary/secondary RDs are $+.023/+.023$, $+.070/+.107$, and $-.167/-.190$ (Table~\ref{tab:rater-sensitivity}). Seven single-label changes among 600 arm labels move Certainty-only$-$Neutral to zero; Confidence$-$Neutral requires 21, and Confidence$-$Existence$\times$Directness requires 50 (Table~\ref{tab:reference-sensitivity}). Under illustrative five-point per-arm prevalence-error budgets, the RD bounds are $[-.077,+.123]$, $[-.030,+.170]$, and $[-.267,-.067]$; only the third sign survives. The budget marks breakpoints, not a ground-truth error rate.

\section{How to decide whether a comparison is supported}
\label{sec:protocol}

Apply the protocol separately to every intended headline.

\paragraph{1. Specify the measurement record.}
Before seeing effects, define the positive label and the unit of analysis, and reproduce both prompts. State the pairing key, target models, items, domains, turns, weights, pipeline and router versions, audit design, and rater process. Prespecify a tolerance $\epsilon$ for automatic--primary-rater contrast bias, and say whether success means preserving the sign or showing equivalence. For each headline, archive the outcome and rubric; prompts and pairing key; target models and weights; pipeline and routing versions; sampling cells and seed; per-arm and per-route confusion counts; same-key effects; dual-rater agreement and adjudication; and the final decision with its prohibited extrapolations.

\paragraph{2. Separate target sampling from failure discovery.}
Draw a probability sample, or an explicitly balanced sample, from both arms. Record inclusion probabilities or target weights. Preserve identical treatment--control keys. A separate hard sample can find failures, but applying its metrics to the target requires a justified model.

\paragraph{3. Construct an LLM-rater process.}
Blind raters to hypotheses, expected directions, and automatic labels. Have both raters independently label a declared subset. Report counts, agreement, $\kappa$, and the adjudication policy. Rater disagreement is outcome uncertainty, not a nuisance to omit.

\paragraph{4. Report errors by arm and route.}
Publish TP/FP/FN/TN, intervals, prevalence, missingness, and routing shares for each arm. For a hybrid scorer, also report within-route behavior and the quantities in Equation~\ref{eq:routing}. Pooled $F_1$ cannot replace this disclosure.

\paragraph{5. Test the comparison on identical cases.}
On the target keys and weights, estimate $\widehat\Delta^{\rm auto}$, $\widehat\Delta^{\rm ref}$, and
$\widehat d=\widehat\Delta^{\rm auto}-\widehat\Delta^{\rm ref}$ with clustering that matches the design. An unpaired slice is only diagnostic.
Here ``ref'' denotes the primary-rater contrast.

\paragraph{6. Apply four gates and a disposition.}
\emph{Construct} asks whether both prompts implement one intended comparison. \emph{Observation} asks whether the audit measured errors by arm. \emph{Applicability (transport)} asks whether the audit applies to $\mathcal P$. \emph{Same-key} asks whether the prespecified contrast rule passes on identical keys. \emph{Retain} when all gates pass. \emph{Qualify} when the stronger rule fails but a prespecified weaker sign rule passes. \emph{Reject} when a sufficiently reliable same-key audit contradicts the raw headline. \emph{Quarantine} when the construct, rater process, or audit cannot answer the claim. Correction is eligible only after the first three gates and cannot repair a changed comparator.

\paragraph{7. Version and revalidate.}
Changes to the prompt, label, routed LLM judge, router, model panel, target weights, sampling, or rater process create a new record. Preserve exploratory status and earlier versions. Do not relabel diagnostics as prospective confirmation.

\paragraph{What an adequate audit costs.}
The paired audit here required $300\times4\times2=2{,}400$ ratings; the representative audit required 288 primary and 96 secondary labels. Planning calculations suggest about 20 primary-rater-negative rows per arm to detect the observed raw extreme, and about 292 for a 10-point difference around $.20$--$.30$ without model clustering. A staged design follows: first size a paired audit to preserve or reject the sign; then add calibration rows only if a correction is needed and transport can be defended; reserve equivalence claims for audits sized to $\epsilon$.

\section{Discussion and conclusion}

The case study shows why trust should attach to a comparison rather than to an evaluator in the abstract. Pooled $F_1=0.916$ coexisted with arm false-positive rates from 0.026 to 0.333, and the difference in those rates was too imprecise to rule out. The stress and balanced audits described different case populations. The alternative automatic judge repeated one ordering but shared rows, framing, and reference labels, so it is corroboration rather than independent validation. On identical cases, one automatic comparison shrank to an interval including zero, and automatic scoring overstated it by 0.070 relative to the primary rater. Two other comparisons kept their direction under both raters, and the prompt-specific correction moved toward the primary-rater estimate but remained too uncertain to replace it.

Validation should therefore attach to a defined outcome, two prompts, a target set of cases, and an audit sample, together with versioned automatic and rater processes. Error counts by prompt explain a problem; same-case ratings test whether the reported difference survives. Neither rescues prompts that measure different constructs or an audit that does not apply to the target population. The protocol uses explicit decisions instead of a universal $F_1$ threshold.

\paragraph{What the case study supports.}
The representative audit supports a local statement: in an equal-model mixture of these six arms, observed FPRs span 0.026--0.333, but the omnibus does not reject equal arm FPRs and modest variation remains unresolved. The alternative judge repeats the Neutral--Existence ordering on the same rows, which is corroboration rather than independent validation. The factorial audit addresses the reported contrasts: Certainty-only$-$Neutral attenuates to an interval spanning zero, Confidence$-$Neutral stays positive under both raters, and Confidence$-$Existence$\times$Directness stays negative under the rater-error budget. These outcomes justify contrast-specific decisions. They do not identify a globally trustworthy scorer or an error budget for future prompts, and arm correction cannot replace the fully rated estimate because calibration is sparse and dependence is only approximated.

The practical lesson is to match evidence to the decision. For a benchmark prevalence claim, estimate error in the benchmark population. For a difference between prompts, rate matched cases and test whether scoring error changes that difference. If the prompts do not measure the same construct, the LLM ratings are unstable, or the audit does not apply to the target cases, quarantine the result instead of turning pooled agreement into confidence. This discipline keeps automatic evaluation useful while making each score's limits explicit.

\section*{Limitations}

The representative audit has 48 responses per arm and eight per model. Precision, FPR, and especially sensitivity intervals are wide. The exploratory omnibus has 102 primary-rater-negative item clusters, treats six models as fixed, and lacks precision for modest arm differences. Factorial intervals are also exploratory.

The audit labels are LLM ratings, not human judgments or ground truth. Both rater families are unused among the evaluated endpoints and automatic judges, but agreement between two LLM families bounds correlated disagreement, not error. A human-labeled slice would let us ask whether the primary rater is the right reference rather than merely a different one. We did not collect one, so every contrast decision in this paper is conditional on the primary LLM rater.

The representative dual subset excludes Spontaneous and Soft cue. Spontaneous has the highest observed false-positive rate, so cross-rater agreement does not constrain the arm where scoring error is largest. Disagreements are not adjudicated, and both raters label all 1,200 factorial responses. The ``representative'' target is only an equal-model mixture of clean nonreferents in six arms. It is not provider traffic, future models, or arbitrary prompts. Hard-sample and route summaries support neither causal nor population claims.

The alternative automatic judge is one correlated frontier model on the same rows. Rogan--Gladen assumes that arm calibration transports to the target and handles overlap only approximately. Its interval rules it out as a replacement headline. The protocol supplies no universal $\epsilon$, and this retrospective study cannot demonstrate prospective preregistration. A decisive demonstration would need a target-matched audit with enough negatives per arm to estimate arm error precisely and a reference label that is not itself an LLM; the present audit is a worked example, not a powered test of arm heterogeneity.

\section*{Data and code availability}

Reproduction materials are available at \url{https://github.com/ancilcrayton/scorer-validity-public}. The release includes the frozen response-level evaluation data (5,428 rows across twelve files), prompts, automatic labels, both LLM-rater label sets, analysis code, and a data card with schemas, hashes, and provenance. No proprietary scorer is claimed.

\section*{Ethical considerations}

Overstated automatic evaluation can mislead benchmarks and model comparisons; arm counts and uncertainty reduce that risk. The data use generated nonreferents and no private conversations, and candidate names were collision-screened against real people. Hosted LLM ratings are neither human labels nor ground truth; no human subjects took part, and model endpoints are not redistributed.

\label{page:last-core}
\clearpage
\bibliographystyle{plainnat}
\bibliography{references}

\clearpage
\appendix

\section{Bias and routing derivations}
\label{app:derivations}

Conditioning on the primary-rater label yields
\[
E[\autoY\mid A=j]
=(1-\beta_j)p_j+\alpha_j(1-p_j).
\]
Subtracting $p_j$ gives $b_j=\alpha_j(1-p_j)-\beta_jp_j$. Subtracting the comparator bias gives Equation~\ref{eq:bias}. Equal $F_1$ values need not imply equal $b_j$: $F_1$ omits true negatives, while signed bias depends on prevalence and both error directions.

Likewise,
\[
E[\autoY-Y\mid A=a]
=\sum_{r\in\{0,1\}}P(R=r\mid A=a)
E[\autoY-Y\mid A=a,R=r],
\]
which gives Equation~\ref{eq:routing}. A pooled hybrid metric can change with routing shares even when behavior within each route is fixed.

\clearpage
\section{Sampling and sensitivity diagnostics}
\label{app:diagnostics}

\begin{table}[H]
\caption{\textbf{Arm precision under hard-enriched versus representative sampling.}
$N$ counts rows labeled by the primary LLM rater (Mistral Large 3 2512). Precision and 95\% Wilson intervals use automatic predicted-positive denominators.
From hard to representative sampling, Neutral rises .667$\rightarrow$.958, Authority rises .633$\rightarrow$.966, and Spontaneous rises .684$\rightarrow$.794.
These shifts depend on the sampling plan; they are not a tested interaction.}
\label{tab:sampling-shift}
\centering
\small
\input{tables/sampling_shift.tex}
\end{table}

The displayed six arms account for 478 of the 550 hard labels; 72 control-layer rows are omitted. We do not inverse-weight the expanded sample because it was tiered by pressure and routed-judge status to find failures.

\begin{table}[H]
\caption{\textbf{Exploratory arm-FPR heterogeneity and audit-size planning.}
The omnibus and 15 pairwise tests use representative rows labeled negative by the primary LLM rater (Mistral Large 3 2512), model fixed effects, and item-clustered CR1 covariance.
Arm heterogeneity is not resolved ($p=.1052$); Soft cue$-$Spontaneous is $-.325$ (Holm $p=.1572$).
Independent planning needs 20 negatives/arm for that raw extreme but 292 for a 10-point difference; these are approximations, not power guarantees.}
\label{tab:heterogeneity-power}
\centering
\small
\input{tables/heterogeneity_power.tex}
\end{table}

\clearpage
\section{Alternative automatic judge details}
\label{app:alternative}

\begin{table}[h!]
\caption{\textbf{Automatic pipeline versus the Claude Sonnet 5 alternative automatic judge on the same 288 representative rows.}
Scorer columns give true positives over predicted positives, with precision in parentheses, against the primary LLM rater (Mistral Large 3 2512). Agreement is also with that rater.
Claude gives Existence precision .842 and Neutral precision 1.000, but it shares rows, framing, and primary-rater labels and scores some Claude outputs.
This is correlated automatic-scoring evidence, not an independent rating study.}
\label{tab:alternative}
\centering
\small
\input{tables/alternative_judge.tex}
\end{table}

\section{LLM-rater sensitivity details}
\label{app:rater-sensitivity}

\begin{table}[h!]
\caption{\textbf{Sensitivity of same-key primary-rater contrasts to label changes.}
The primary LLM rater is Mistral Large 3 2512. ``Flips'' is the minimum number of adversarial changes among 600 arm labels that erases the primary-rater RD sign. Bounds allow $\pm.05$ prevalence error per arm.
Certainty-only$-$Neutral is erased by seven flips, whereas only Confidence$-$Existence$\times$Directness retains its sign under the five-point budget [$-.267,-.067$].
The budget is illustrative and is close to, but not estimated from, the observed 4.75\% cross-rater disagreement.}
\label{tab:reference-sensitivity}
\centering
\small
\input{tables/reference_sensitivity.tex}
\end{table}

\begin{table}[h!]
\caption{\textbf{Full factorial sensitivity across unused-family LLM raters.}
Mistral Large 3 2512 and MiniMax M3 independently label all 1,200 responses on the same 300 keys.
Their RDs differ by at most 3.7 points across the three contrasts; neither rater supplies ground truth.}
\label{tab:rater-sensitivity}
\centering
\small
\input{tables/rater_sensitivity.tex}
\end{table}

\clearpage
\section{Routing details}
\label{app:routing}

\begin{table}[h!]
\caption{\textbf{Route composition and primary-rater-relative performance in the 550-row hard audit.}
The primary LLM rater is Mistral Large 3 2512, and the routed LLM judge is DeepSeek V4 Pro. Rows report sample share, confusion counts, precision, recall, and $F_1$ for heuristic-only and routed-judge cases.
Routed cases are 48.0\% with $F_1=.545$, versus .762 for heuristic-only cases; pooled $F_1=.669$.
Harder responses are more likely to be routed, so this difference is descriptive and not causal.}
\label{tab:routing}
\centering
\small
\input{tables/hard_audit_routing.tex}
\end{table}

\section{Existence error taxonomy}
\label{app:existence-taxonomy}

The automatic pipeline has three Existence false positives, all hypothetical abstentions or requests for context. The alternative automatic judge also has three, all hypothetical or speculative completions. We assigned categories after reading the response text. The analysis is descriptive, used one coder, and does not provide another LLM rating.

\begin{table}[h!]
\caption{\textbf{Exploratory taxonomy of primary-rater-relative Existence false positives on identical representative rows.}
The primary LLM rater is Mistral Large 3 2512. Exclusive categories compare the automatic pipeline with the Claude Sonnet 5 alternative automatic judge.
All three pipeline false positives are hypothetical abstentions/context requests; all three alternative-judge false positives are hypothetical/speculative completions.
Categories were single-coded post hoc.}
\label{tab:existence-taxonomy}
\centering
\small
\input{tables/existence_fp_taxonomy.tex}
\end{table}

\section{Secondary transfer cases}
\label{app:external}

These cases illustrate the protocol; they do not evaluate any official benchmark scorer. Both concern our non-original, task-specific implementations, and neither decision extends to published systems.

\paragraph{GaslightBench.}
Our incorrectness heuristic has TP/FP/FN/TN $=77/132/15/76$ on $n=300$ primary-rater judgments: precision $.368$, recall $.837$, and $F_1=.512$. Arm $F_1$ ranges from $.113$ on Neutral T2 to $.927$ on Existence T2. Dual-rater $\kappa=.573$ does not repair the heuristic's disagreement with the primary LLM rater. We quarantine automatic effects rather than replace them with a primary-rater headline. Without same-key automatic--primary-rater intervals, the audit cannot reject a particular contrast.

\paragraph{SYCON.}
Our non-original automatic acceptance judge has TP/FP/FN/TN $=332/21/320/227$ on $n=900$ turn judgments ($F_1=.661$; dual $\kappa=.635$). Neutral recall is $65/183=.355$, versus $170/246=.691$ for Existence. In 60 audited topic--model trajectories, primary-rater ever-accept is Existence $60/60=1.000$ (Wilson $[.940,1.000]$), pressure $55/60=.917$ ($[.819,.964]$), and Neutral $49/60=.817$ ($[.701,.894]$). We retain only this descriptive ordering. Existence was not the published no-pressure control, the intervals are marginal rather than paired, and we claim no reversal of the published conclusion \citep{hong-etal-2025-measuring}. We likewise cite the Gaslight task only as the source setting \citep{cui-etal-2025-gaslightbench}.

\begin{table}[H]
\caption{\textbf{Transfer-audit dispositions for two non-original scorers.}
$N$ counts turn judgments labeled by the primary LLM rater (Mistral Large 3 2512). Confusion counts and $F_1$ compare each automatic implementation with those labels.
Gaslight has $F_1=.512$ and is quarantined. The SYCON automatic judge has $F_1=.661$ and supports only a qualified descriptive ordering.
Neither decision applies beyond the analyzed label and sample.}
\label{tab:external}
\centering
\small
\input{tables/external_cases.tex}
\end{table}

\clearpage
\section{Claims withheld by the contrast-level validity gate}
\label{app:card}

\begin{table}[H]
\caption{\textbf{Claims withheld by the contrast-level validity gate.}
Authority$-$Existence is quarantined because its prompt permits hypothetical completion, despite Existence precision .833. Gaslight lacks same-key intervals. We claim no SYCON reversal because its automatic judge is non-original and Neutral recall is .355.
Reviving these claims requires a construct-compatible comparator and/or a target-matched audit, not pooled $F_1$.}
\label{tab:quarantine}
\centering
\small
\input{tables/quarantine_register.tex}
\end{table}

\end{document}

%% file: tables/representative_audit.tex
\begin{tabular}{@{}lcccc@{}}
\toprule
\textbf{Arm} & \textbf{TP/FP/FN/TN} & \textbf{Precision [95\% CI]} & \textbf{FPR [95\% CI]} & \textbf{$\widehat b_j$} \\
\midrule
Spontaneous & 27/7/0/14 & .794 [.632,.897] & .333 [.172,.546] & +.146 \\
Soft cue & 7/1/3/37 & .875 [.529,.978] & .026 [.005,.135] & $-.042$ \\
Neutral & 23/1/0/24 & .958 [.798,.993] & .040 [.007,.195] & +.021 \\
Existence & 15/3/1/29 & .833 [.608,.942] & .094 [.032,.242] & +.042 \\
Confidence & 20/5/0/23 & .800 [.609,.911] & .179 [.079,.356] & +.104 \\
Authority & 28/1/0/19 & .966 [.828,.994] & .050 [.009,.236] & +.021 \\
\bottomrule
\end{tabular}

%% file: tables/factorial_same_key.tex
\begin{tabular}{@{}lcccc@{}}
\toprule
\textbf{Contrast} & \textbf{\shortstack{Auto\\RD}} & \textbf{\shortstack{Primary-rater\\RD [95\% CI]}} & \textbf{\shortstack{Auto $-$ primary\\{[95\% CI]}}} & \textbf{Decision} \\
\midrule
Cert-only $-$ Neutral & +.093 & $+.023$ [$-.032$,$+.078$] & $+.070$ [$+.013$,$+.128$] & spans 0; auto overstates \\
Conf $-$ Exist$\times$Dir & $-.163$ & $-.167$ [$-.223$,$-.111$] & $+.003$ [$-.033$,$+.040$] & sign retained \\
Conf $-$ Neutral & +.113 & $+.070$ [$+.021$,$+.118$] & $+.043$ [$+.003$,$+.085$] & sign retained \\
\bottomrule
\end{tabular}

%% file: tables/arm_correction.tex
\begin{tabular}{@{}lrrrr@{}}
\toprule
\textbf{Arm/contrast} & \textbf{Auto} & \textbf{Primary} & \textbf{Audit Se/Sp} & \textbf{Corrected [95\%]} \\
\midrule
Neutral & 0.397 & 0.400 & 1.000/0.960 & 0.372 [0.261,0.449] \\
Confidence & 0.510 & 0.470 & 1.000/0.821 & 0.403 [0.240,0.523] \\
\midrule
Confidence $-$ Neutral & +0.113 & +0.070 & --- & +0.032 [-0.146,+0.197] \\
\bottomrule
\end{tabular}

%% file: tables/sampling_shift.tex
\begin{tabular}{@{}lrrrr@{}}
\toprule
\textbf{Arm} & \textbf{Hard $N$} & \textbf{Hard precision [CI]} & \textbf{Rep.\ $N$} & \textbf{Rep.\ precision [CI]} \\
\midrule
Spontaneous & 24 & .684 [.460,.846] & 48 & .794 [.632,.897] \\
Soft cue & 74 & .800 [.584,.919] & 48 & .875 [.529,.978] \\
Neutral & 81 & .667 [.454,.828] & 48 & .958 [.798,.993] \\
Existence & 63 & .759 [.579,.878] & 48 & .833 [.608,.942] \\
Confidence & 125 & .625 [.453,.771] & 48 & .800 [.609,.911] \\
Authority & 111 & .633 [.455,.781] & 48 & .966 [.828,.994] \\
\bottomrule
\end{tabular}

%% file: tables/heterogeneity_power.tex
\begin{tabular}{@{}>{\raggedright\arraybackslash}p{.31\columnwidth}>{\raggedright\arraybackslash}p{.62\columnwidth}@{}}
\toprule
\textbf{Diagnostic} & \textbf{Result and interpretation} \\
\midrule
Arm omnibus & $\chi^2_{5}=9.10$, $p=.1052$; model fixed effects, clustered by item. \\
Strongest pair & Soft cue $-$ Spontaneous: $-.325$ (Holm $p=.1572$); exploratory, 15 pairs. \\
Observed-extreme planning & 20 negatives/arm; independent approximation. \\
10-point planning & 292 negatives/arm; independent approximation. \\
\bottomrule
\end{tabular}

%% file: tables/alternative_judge.tex
\begin{tabular}{@{}lccc@{}}
\toprule
\textbf{Arm} & \textbf{\shortstack{Pipeline\\TP/auto$+$ ($P$)}} & \textbf{\shortstack{Claude auto\\TP/auto$+$ ($P$)}} & \textbf{\shortstack{Claude--primary\\agreement}} \\
\midrule
Spontaneous & 27/34 (.794) & 25/26 (.962) & .938 \\
Soft cue & 7/8 (.875) & 9/10 (.900) & .958 \\
Neutral & 23/24 (.958) & 23/23 (1.000) & 1.000 \\
Existence & 15/18 (.833) & 16/19 (.842) & .938 \\
Confidence & 20/25 (.800) & 18/21 (.857) & .896 \\
Authority & 28/29 (.966) & 28/30 (.933) & .958 \\
\bottomrule
\end{tabular}

%% file: tables/reference_sensitivity.tex
\begin{tabular}{@{}>{\raggedright\arraybackslash}p{.38\columnwidth}>{\centering\arraybackslash}p{.25\columnwidth}>{\centering\arraybackslash}p{.27\columnwidth}@{}}
\toprule
\textbf{Contrast} & \textbf{Primary RD / flips (\% labels)} & \textbf{RD bound, $\epsilon=.05$} \\
\midrule
Certainty-only $-$ Neutral & $+.023$ / 7 (1.2) & [$-.077$,$+.123$] \\
Confidence $-$ Existence$\times$Directness & $-.167$ / 50 (8.3) & [$-.267$,$-.067$] \\
Confidence $-$ Neutral & $+.070$ / 21 (3.5) & [$-.030$,$+.170$] \\
\bottomrule
\end{tabular}

%% file: tables/rater_sensitivity.tex
\begin{tabular}{@{}>{\raggedright\arraybackslash}p{.40\columnwidth}>{\centering\arraybackslash}p{.30\columnwidth}>{\centering\arraybackslash}p{.20\columnwidth}@{}}
\toprule
\textbf{Contrast} & \textbf{Primary / secondary RD} & \textbf{Difference} \\
\midrule
Certainty-only $-$ Neutral & $+.023$ / $+.023$ & $+.000$ \\
Confidence $-$ Existence$\times$Directness & $-.167$ / $-.190$ & $-.023$ \\
Confidence $-$ Neutral & $+.070$ / $+.107$ & $+.037$ \\
\bottomrule
\end{tabular}

%% file: tables/hard_audit_routing.tex
\begin{tabular}{@{}lrrrrr@{}}
\toprule
\textbf{Route} & \textbf{$N$ (share)} & \textbf{TP/FP/FN/TN} & \textbf{$P$} & \textbf{$R$} & \textbf{$F_1$} \\
\midrule
Heuristic-only & 286 (52.0\%) & 72/28/17/169 & .720 & .809 & .762 \\
Routed LLM judge & 264 (48.0\%) & 39/24/41/160 & .619 & .488 & .545 \\
\midrule
Pooled & 550 (100\%) & 111/52/58/329 & .681 & .657 & .669 \\
\bottomrule
\end{tabular}

%% file: tables/existence_fp_taxonomy.tex
\begin{tabular}{@{}lrr@{}}
\toprule
\textbf{Exclusive response category} & \textbf{Pipeline FP} & \textbf{Claude FP} \\
\midrule
Hypothetical/speculative completion & 0 & 3 \\
Hypothetical abstention / needs context & 3 & 0 \\
Referent substitution & 0 & 0 \\
\midrule
\textbf{Total Existence false positives} & \textbf{3} & \textbf{3} \\
\bottomrule
\end{tabular}

%% file: tables/external_cases.tex
\begin{tabular}{@{}lrrrl@{}}
\toprule
\textbf{Case} & \textbf{$N$} & \textbf{TP/FP/FN/TN} & \textbf{$F_1$} & \textbf{Decision} \\
\midrule
Gaslight heuristic & 300 & 77/132/15/76 & .512 & quarantine effects \\
SYCON & 900 & 332/21/320/227 & .661 & qualify ordering only \\
\bottomrule
\end{tabular}

%% file: tables/quarantine_register.tex
\begin{tabular}{@{}>{\raggedright\arraybackslash}p{.29\columnwidth}>{\raggedright\arraybackslash}p{.64\columnwidth}@{}}
\toprule
\textbf{Candidate claim} & \textbf{Disposition and binding validity problem} \\
\midrule
H3 Auth.--Exist. & \textbf{Quarantined.} The comparator licenses hypothetical completion, so it changes the construct despite representative precision 15/18=.833 [.608,.942]. \\
Our Gaslight heuristic effects & \textbf{Quarantined.} Poor overall audit; no same-key automatic--primary-rater contrast intervals. \\
SYCON reversal & \textbf{No reversal claim.} Non-original automatic judge; Neutral recall .355 ($n=300$ judgments). \\
\bottomrule
\end{tabular}

%% file: references.bib
@inproceedings{min-etal-2023-factscore,
  title = {{FActScore}: Fine-grained Atomic Evaluation of Factual Precision in Long Form Text Generation},
  author = {Min, Sewon and Krishna, Kalpesh and Lyu, Xinxi and Lewis, Mike and Yih, Wen-tau and Koh, Pang and Iyyer, Mohit and Zettlemoyer, Luke and Hajishirzi, Hannaneh},
  booktitle = {Proceedings of the 2023 Conference on Empirical Methods in Natural Language Processing},
  year = {2023},
  address = {Singapore},
  publisher = {Association for Computational Linguistics},
  pages = {12076--12100},
  doi = {10.18653/v1/2023.emnlp-main.741},
  url = {https://aclanthology.org/2023.emnlp-main.741/}
}

@inproceedings{yona-etal-2024-large,
  title = {Can Large Language Models Faithfully Express Their Intrinsic Uncertainty in Words?},
  author = {Yona, Gal and Aharoni, Roee and Geva, Mor},
  booktitle = {Proceedings of the 2024 Conference on Empirical Methods in Natural Language Processing},
  year = {2024},
  address = {Miami, Florida, USA},
  publisher = {Association for Computational Linguistics},
  pages = {7752--7764},
  doi = {10.18653/v1/2024.emnlp-main.443},
  url = {https://aclanthology.org/2024.emnlp-main.443/}
}

@inproceedings{madhusudhan-etal-2025-llms,
  title = {Do {LLM}s Know When to {NOT} Answer? Investigating Abstention Abilities of Large Language Models},
  author = {Madhusudhan, Nishanth and Madhusudhan, Sathwik Tejaswi and Yadav, Vikas and Hashemi, Masoud},
  booktitle = {Proceedings of the 31st International Conference on Computational Linguistics},
  year = {2025},
  address = {Abu Dhabi, UAE},
  publisher = {Association for Computational Linguistics},
  pages = {9329--9345},
  url = {https://aclanthology.org/2025.coling-main.627/}
}

@inproceedings{hong-etal-2025-measuring,
  title = {Measuring Sycophancy of Language Models in Multi-turn Dialogues},
  author = {Hong, Jiseung and Byun, Grace and Kim, Seungone and Shu, Kai},
  booktitle = {Findings of the Association for Computational Linguistics: EMNLP 2025},
  year = {2025},
  address = {Suzhou, China},
  publisher = {Association for Computational Linguistics},
  pages = {2239--2259},
  doi = {10.18653/v1/2025.findings-emnlp.121},
  url = {https://aclanthology.org/2025.findings-emnlp.121/}
}

@inproceedings{cui-etal-2025-gaslightbench,
  title = {{GASLIGHTBENCH}: Quantifying {LLM} Susceptibility to Social Prompting},
  author = {Yao, Xuanzhe and Ghosh, Sahil and Lee, Gareth and Logian, William H. and Cui, Lening Nick and Podoshev, Ellie and Srivastava, Swarit and Li, Michael and Sandoval, Aaron and O'Brien, Sean and Saxon, Michael and Dev, Sunishchal and Zhu, Kevin},
  booktitle = {NeurIPS 2025 Workshop on Evaluating the Evolving LLM Lifecycle: Benchmarks, Emergent Abilities, and Scaling},
  year = {2025},
  url = {https://openreview.net/forum?id=0BYRYwGCbK},
  note = {Equal contribution among Yao, Ghosh, Lee, Logian, and Cui}
}

@inproceedings{liu-etal-2023-g,
  title = {{G}-Eval: {NLG} Evaluation using {GPT}-4 with Better Human Alignment},
  author = {Liu, Yang and Iter, Dan and Xu, Yichong and Wang, Shuohang and Xu, Ruochen and Zhu, Chenguang},
  booktitle = {Proceedings of the 2023 Conference on Empirical Methods in Natural Language Processing},
  year = {2023},
  pages = {2511--2522},
  publisher = {Association for Computational Linguistics},
  doi = {10.18653/v1/2023.emnlp-main.153},
  url = {https://aclanthology.org/2023.emnlp-main.153/}
}

@inproceedings{zheng-etal-2023-llm-judge,
  title = {Judging {LLM}-as-a-Judge with {MT}-Bench and Chatbot Arena},
  author = {Zheng, Lianmin and Chiang, Wei-Lin and Sheng, Ying and Zhuang, Siyuan and Wu, Zhanghao and Zhuang, Yonghao and Lin, Zi and Li, Zhuohan and Li, Dacheng and Xing, Eric P. and Zhang, Hao and Gonzalez, Joseph E. and Stoica, Ion},
  booktitle = {Advances in Neural Information Processing Systems},
  year = {2023},
  url = {https://openreview.net/forum?id=uccHPGDlao}
}

@article{angelopoulos-etal-2023-prediction,
  title = {Prediction-Powered Inference},
  author = {Angelopoulos, Anastasios N. and Bates, Stephen and Fannjiang, Clara and Jordan, Michael I. and Zrnic, Tijana},
  journal = {Science},
  volume = {382},
  number = {6671},
  pages = {669--674},
  year = {2023},
  doi = {10.1126/science.adi6000}
}

@article{rogan-gladen-1978-prevalence,
  title = {Estimating Prevalence from the Results of a Screening Test},
  author = {Rogan, Walter J. and Gladen, Beth C.},
  journal = {American Journal of Epidemiology},
  volume = {107},
  number = {1},
  pages = {71--76},
  year = {1978},
  doi = {10.1093/oxfordjournals.aje.a112510}
}

@inproceedings{wang-etal-2024-fair-evaluators,
  title = {Large Language Models are not Fair Evaluators},
  author = {Wang, Peiyi and Li, Lei and Chen, Liang and Cai, Zefan and Zhu, Dawei and Lin, Binghuai and Cao, Yunbo and Kong, Lingpeng and Liu, Qingfu and Liu, Tianyu and Sui, Zhifang},
  booktitle = {Proceedings of the 62nd Annual Meeting of the Association for Computational Linguistics (Volume 1: Long Papers)},
  year = {2024},
  pages = {9440--9450},
  publisher = {Association for Computational Linguistics},
  doi = {10.18653/v1/2024.acl-long.511},
  url = {https://aclanthology.org/2024.acl-long.511/}
}

@inproceedings{panickssery-etal-2024-evaluators,
  title = {{LLM} Evaluators Recognize and Favor Their Own Generations},
  author = {Panickssery, Arjun and Bowman, Samuel R. and Feng, Shi},
  booktitle = {Advances in Neural Information Processing Systems},
  year = {2024},
  volume = {37},
  url = {https://arxiv.org/abs/2404.13076}
}

@inproceedings{bavaresco-etal-2025-llm-judges,
  title = {{LLM}s instead of Human Judges? A Large Scale Empirical Study across 20 {NLP} Evaluation Tasks},
  author = {Bavaresco, Anna and Bernardi, Raffaella and Bertolazzi, Leonardo and Elliott, Desmond and Fern{\'a}ndez, Raquel and Gatt, Albert and Ghaleb, Esam and Giulianelli, Mario and others},
  booktitle = {Proceedings of the 63rd Annual Meeting of the Association for Computational Linguistics},
  year = {2025},
  publisher = {Association for Computational Linguistics},
  url = {https://arxiv.org/abs/2406.18403}
}

@inproceedings{sclar-etal-2024-format-spread,
  title = {Quantifying Language Models' Sensitivity to Spurious Features in Prompt Design or: How {I} learned to start worrying about prompt formatting},
  author = {Sclar, Melanie and Choi, Yejin and Tsvetkov, Yulia and Suhr, Alane},
  booktitle = {International Conference on Learning Representations},
  year = {2024},
  url = {https://arxiv.org/abs/2310.11324}
}

@inproceedings{jacobs-wallach-2021-measurement,
  title = {Measurement and Fairness},
  author = {Jacobs, Abigail Z. and Wallach, Hanna},
  booktitle = {Proceedings of the 2021 ACM Conference on Fairness, Accountability, and Transparency},
  year = {2021},
  pages = {375--385},
  publisher = {ACM},
  doi = {10.1145/3442188.3445901}
}

@inproceedings{raji-etal-2021-everything,
  title = {{AI} and the Everything in the Whole Wide World Benchmark},
  author = {Raji, Inioluwa Deborah and Denton, Emily and Bender, Emily M. and Hanna, Alex and Paullada, Amandalynne},
  booktitle = {Proceedings of the NeurIPS Track on Datasets and Benchmarks},
  year = {2021},
  volume = {1},
  url = {https://arxiv.org/abs/2111.15366}
}

@article{bross-1954-misclassification,
  title = {Misclassification in 2 {X} 2 Tables},
  author = {Bross, Irwin},
  journal = {Biometrics},
  volume = {10},
  number = {4},
  pages = {478--486},
  year = {1954},
  doi = {10.2307/3001619}
}

@article{copeland-etal-1977-bias,
  title = {Bias Due to Misclassification in the Estimation of Relative Risk},
  author = {Copeland, Kenneth T. and Checkoway, Harvey and McMichael, Anthony J. and Holbrook, Robert H.},
  journal = {American Journal of Epidemiology},
  volume = {105},
  number = {5},
  pages = {488--495},
  year = {1977},
  doi = {10.1093/oxfordjournals.aje.a112408}
}

@article{greenland-1980-effect,
  title = {The Effect of Misclassification in the Presence of Covariates},
  author = {Greenland, Sander},
  journal = {American Journal of Epidemiology},
  volume = {112},
  number = {4},
  pages = {564--569},
  year = {1980},
  doi = {10.1093/oxfordjournals.aje.a113025}
}

@inproceedings{pearl-bareinboim-2011-transportability,
  title = {Transportability of Causal and Statistical Relations: A Formal Approach},
  author = {Pearl, Judea and Bareinboim, Elias},
  booktitle = {Proceedings of the 25th AAAI Conference on Artificial Intelligence},
  year = {2011},
  pages = {247--254},
  doi = {10.1609/aaai.v25i1.7861}
}

@inproceedings{egami-etal-2023-imperfect,
  title = {Using Imperfect Surrogates for Downstream Inference: Design-based Supervised Learning for Social Science Applications of Large Language Models},
  author = {Egami, Naoki and Hinck, Musashi and Stewart, Brandon M. and Wei, Hanying},
  booktitle = {Advances in Neural Information Processing Systems},
  year = {2023},
  volume = {36},
  url = {https://arxiv.org/abs/2306.04746}
}

@inproceedings{thakur-etal-2025-judging,
  title = {Judging the Judges: Evaluating Alignment and Vulnerabilities in {LLM}s-as-Judges},
  author = {Thakur, Aman Singh and Choudhary, Kartik and Ramayapally, Venkat Srinik and Vaidyanathan, Sankaran and Hupkes, Dieuwke},
  booktitle = {Proceedings of the 2nd Workshop on Generation, Evaluation and Metrics (GEM$^2$)},
  year = {2025},
  pages = {404--430},
  url = {https://arxiv.org/abs/2406.12624}
}
